%% file: PreViTS.tex
\documentclass[10pt,twocolumn,letterpaper]{article}

\usepackage{cvpr}              

\usepackage{graphicx}
\usepackage{amsmath}
\usepackage{amssymb}
\usepackage{booktabs}

\usepackage[pagebackref,breaklinks,colorlinks]{hyperref}

\usepackage[capitalize]{cleveref}
\crefname{section}{Sec.}{Secs.}
\Crefname{section}{Section}{Sections}
\Crefname{table}{Table}{Tables}
\crefname{table}{Tab.}{Tabs.}

\usepackage{algorithm}
\usepackage{amsmath}
\usepackage{amssymb}
\usepackage[toc,page]{appendix}
\usepackage{booktabs}
\usepackage{color}
\usepackage{colortbl}
\usepackage{comment}
\usepackage{enumitem}
\usepackage{epsfig}
\usepackage{etoolbox}
\usepackage{graphicx}
\usepackage[utf8]{inputenc}
\usepackage{listings}
\usepackage{multirow}
\usepackage[numbers,sort&compress]{natbib}
\usepackage[olditem,oldenum]{paralist}
\usepackage{pifont}
\usepackage{tabularx}
\usepackage{times}
\usepackage{transparent}
\usepackage[dvipsnames]{xcolor}
\usepackage{makecell}
\usepackage{subcaption}
\usepackage{tabularx}
\input{definitions.tex}

\usepackage{inconsolata}

\usepackage{cleveref}

\newcommand{\app}{PreViTS}

\newcommand{\fullapp}{Pretraining with Video Tracking Supervision}

\definecolor{orange}{rgb}{1,0.5,0}
\definecolor{lightsalmonpink}{rgb}{1.0, 0.6, 0.6}
\definecolor{verylightsalmonpink}{rgb}{0.966, 0.805, 0.797}
\definecolor{lightblue}{rgb}{0.862, 0.906, 0.984}
\definecolor{lightyellow}{rgb}{1.0, 0.945, 0.797}
\definecolor{lightgreen}{rgb}{0.835, 0.91, 0.828}
\definecolor{lightpurple}{rgb}{0.879, 0.832, 0.902}

\newcommand{\bccomment}[1]{\textcolor{black}{ #1}}

\newcommand{\imagenet}[0]{ImageNet-1k}
\newcommand{\voc}[0]{PASCAL VOC}
\newcommand{\inclf}[0]{IN-1k}
\newcommand{\ucf}[0]{UCF-101}
\newcommand{\vocclf}[0]{VOC07}

\newcommand{\random}[0]{Random Init}

\newcommand{\ttbf}[1]{\textbf{\texttt{#1}}}
\newcommand{\graycell}{\cellcolor{gray!20}}
\newcommand{\band}{\rowcolor{gray!20}}
\newcommand{\drop}[1]{\textcolor{gray}{\textsubscript{$-$#1}}}
\newcommand{\rise}[1]{\textcolor{gray}{\textsubscript{$+$#1}}}
\newcommand{\riseneg}[1]{\textcolor{gray}{\textsubscript{$-$#1}}}

\newcommand{\Drop}[1]{\textcolor{Red}{\textsubscript{\bf $-$#1}}}
\newcommand{\Rise}[1]{\textcolor{Green}{\textsubscript{\bf $+$#1}}}
\newcommand{\Riseneg}[1]{\textcolor{Green}{\textsubscript{\bf $-$#1}}}

\def\cvprPaperID{454} 
\def\confName{CVPR}
\def\confYear{2022}

\begin{document}

\title{PreViTS: Contrastive Pretraining with Video Tracking Supervision}

\author{%
    Brian Chen $^1$  \quad
    Ramprasaath R. Selvaraju $^2$  \quad
    Shih-Fu Chang$^{1}$ \quad
    Juan Carlos Niebles $^2$  \quad
    Nikhil Naik $^2$  
    \vspace{1mm} \\
    \small{$^1$Columbia University, $^2$artera.ai, $^2$Salesforce Research},  \\
    \small{
    \texttt{\{bc2754,sc250\}@columbia.edu},\texttt{ram@artera.ai},\texttt{\{rselvaraju,jniebles,nnaik\}@salesforce.com}} 
}

\maketitle

\input{sections/0_abstract.tex}
\input{sections/1_introduction.tex}

\input{sections/2_related_work.tex}
\input{sections/3_method.tex}
\input{sections/4_experiments.tex}

\input{sections/5_conclusion.tex}

\clearpage

{\small
\bibliographystyle{ieee_fullname}
\bibliography{egbib}
}

\clearpage

\numberwithin{equation}{section}
\setcounter{table}{0}
\setcounter{figure}{0}
\setcounter{equation}{0}

\clearpage 
\setcounter{section}{0}

\noindent \textbf{This appendix is organized as follows:} \\
1. Implementation details. \\
2. Additional experimental results. \\
3. Experiment details. \\
4. Additional qualitative results.



\section{Implementation details}
Image model is from MoCo, video model is from RSPNet.
For experiments with the image model, we use the ResNet-50 backbone and sample one frame with 224 $\times$ 224 spatial sizes for each clip. For experiments with the video model, we use an S3D-g \cite{xie2018rethinking} backbone and sample 16 continuous frames with 224 $\times$ 224 spatial sizes for each clip. We perform standard data augmentation on clips, including random Gaussian blur, and random color jitter~\cite{chen2020simple}. 
To compare with other baseline methods, we also trained on R(2+1)D\cite{tran2018closer}, and C3D\cite{tran2015learning} backbone following \cite{chen2021rspnet}.
We followed \cite{chen2021rspnet} to train our model with 200 epochs with SGD and a batch size of 256. We apply a cosine learning rate scheduler with an LR of 0.03 for the image model and 0.5 for the video model.
Following He~\etal\cite{he2019moco}, we set $\tau = 0.07$, $K = 65535$, $\gamma = 0.15$, $\mu = 0.3$, $\lambda=3$. 
The training time is two days for pretraining VGG-Sound and three days for pretraining on Kinetics. 
For both image and video tasks, we compare with the following baselines: (1) \textbf{\random{}} of weights without pretraining, (2) \textbf{MoCo/RSPNet} to demonstrate standard self-supervised model performance for image (MoCo) and video (RSPNet), (3) \textbf{MoCo/RSPNet + Tracking Constrained Sampling} to evaluate our unsupervised tracking-based spatial-temporal sampling strategy. 

\section{Additional experimental results}

\input{tables/image_ex}

\noindent\textbf{Generalize to image recognition tasks.}
\label{sec:downstream_eval}
We evaluate our learned features on four downstream image recognition tasks: \textbf{(a)} \voc{}~\cite{everingham2009voc} linear classification, \textbf{(b)} \imagenet{}~\cite{deng2009imagenet,russakovsky2015imagenet} linear classification, \textbf{(c)} \voc{} object detection, and \textbf{(d)} COCO~\cite{lin2014microsoft} instance segmentation.
Following \cite{desai2020virtex,selvaraju2021casting}, for \textbf{(a, b)}, we perform linear classification by using the SSL model as a frozen feature extractor and training a classifier on top. For \textbf{(c, d)}, we use the SSL model as weight
initialization for fine-tuning on the labeled datasets. Detailed experimental settings can be found in the supplementary.
Our results in \Cref{tab:main_results} show that training \app{} outperforms baseline MoCo training on all tasks, obtaining robust gains in VOC and ImageNet classification, along with VOC detection and COCO instance classification. Notably, the performance gains when pretraining on VGG-Sound are larger as compared to those on Kinetics-400, even though Kinetics-400 is 20\% larger in terms of the number of videos. 
We speculate that due to VGG-Sound containing a more diverse collection of objects as compared to Kinetics-400, which is primarily human action-centric, VGG-Sound benefits more from being able to learn object-focused representations when training with \app{}. 
The performance improvement over baseline is especially large on the VOC detection task, aided by the improved ability to localize objects during pretraining. Finally, while it is typically challenging to obtain comparable performance to supervised ImageNet pretraining using video SSL pretraining on image recognition tasks \cite{purushwalkam2020demystifying}, due to the larger domain shift, MoCo models trained with \app{} still obtain comparable or better performance to ImageNet-fully supervised training on VOC detection and COCO instance segmentation tasks.

\noindent\textbf{Video Backgrounds Challenge (mini-Kinetics).}
In addition to the video backgrounds challenge, we also evaluate robustness to background signal on the mini-Kinetics dataset~\cite{danceinmall}, a subset of  Kinetics-400  designed to study if video classification models depend on the background signal for scene classification. This dataset contains foreground bounding boxes computed by a person detection model. We utilize the bounding boxes to mask the foreground object to analyze if the model depended on scene features when performing action classification. The model with \app{} achieved an accuracy of 55.24\% in the Original setting compared to 47.18\% for the baseline RSPNet.  When the foreground was masked (No-FG), the accuracy for \app{} drops by 6.9\%, as compared to a drop of 2.71\% for the baseline model, indicating that the \app{}-trained model relies less on the background signal.
\input{sup/percentage}

\noindent\textbf{Computational resource compared to baseline.~}
Obtaining tracking for a dataset is a fixed, one-time computational cost. During training, \app{} only needs 1.3x GPU memory and training time due to the extra forward pass for the foreground key and query to compute Grad-CAM. \app{} is also efficient, it outperforms baseline with only half of the training data (VGG-Sound), i.e., 65\% of its training time in Figure \ref{fig:percentage}.

\noindent\textbf{Method Complexity of \app{}.~}
While \app{} contains several components, it is not sensitive to their hyperparameters and design choices. To test sensitivity, we randomly chose a combination of parameters $\mu$, $\lambda$, using the setting in Tab. 1(8) in the main paper and obtained \textcolor{ForestGreen}{\textbf{+4.32}} \vocclf{} mAP over the baseline, only lower by \textcolor{gray}{\textbf{-0.38}}  than our best model.

\noindent\textbf{Evidence for lack of proper supervisory signal in current SSL approaches.~}
As visualized in Fig. 1(d) in the main paper, simply applying contrastive loss may lead to learning background correlation when the backgrounds are similar. 
Moreover, through a study using supervised segmentation on VGGSound, we found that traditional SSL approaches sample different concepts as positive pairs 27\% of the time, while only 7\% with our spatio-temporal sampling strategy. 
This indicates our strategy can acquire a cleaner supervisory signal. 

\section{Experiment details}

\input{figures/image_back}
\noindent\textbf{Image Backgrounds Challenge.} The settings of different scenarios of backgrounds are shown in Figure \ref{fig:backgrounds}. The figure is from \cite{xiao2020noise}.

\noindent\textbf{Code of the paper.} We will release our code by the time when the paper is published. 

\input{sup/sup_ucf}
\input{sup/sup_bg}

\section{Additional qualitative results}
We include more visualizations for UCF-101 action recognition in Figure \ref{fig:sup_ucf}, Video Backgrounds Challenge in Figure \ref{fig:sup_bg}, and DAVIS video object segmentation in Figure \ref{fig:sup_davis} and \ref{fig:sup_davis2}. 


\input{sup/sup_davis}
\input{sup/sup_davis2}

\end{document}

%% file: definitions.tex
\newcommand{\myapprox}{{\raise.17ex\hbox{$\scriptstyle\sim$}}}

\newcommand{\reffig}[1]{Figure~\ref{#1}}

\def\etal{\textit{et al}.~}

\def\mL{{\mathcal L}}

\DeclareMathAlphabet\mathbfcal{OMS}{cmsy}{b}{n}

\def\0{{\bf 0}}
\usepackage{ dsfont }
\def\1{\mathds{1}}

\usepackage{ntheorem}

\newtheorem*{*thm}{Theorem}

\newtheorem*{*lemma}{Lemma}

\usepackage{enumitem}

%% file: sections/0_abstract.tex
\begin{abstract}
Videos are a rich source for self-supervised learning (SSL) of visual representations due to the presence of natural temporal transformations of objects. However, current methods typically randomly sample video clips for learning, which results in an imperfect supervisory signal. In this work, we propose PreViTS, an SSL framework that utilizes an unsupervised tracking signal for selecting clips containing the same object, which helps better utilize temporal transformations of objects. PreViTS further uses the tracking signal to spatially constrain the frame regions to learn from and trains the model to locate meaningful objects by providing supervision on Grad-CAM attention maps.
To evaluate our approach, we train a momentum contrastive (MoCo) encoder on VGG-Sound and Kinetics-400 datasets with PreViTS. Training with PreViTS outperforms representations learnt by contrastive strategy alone on video downstream tasks, obtaining state-of-the-art performance on action classification. PreViTS helps learn feature representations that are more robust to changes in background and context, as seen by experiments on datasets with background changes. Our experiment also demonstrates various visual transformation invariance captured by our model. Learning from large-scale videos with PreViTS could lead to more accurate and robust visual feature representations. 
\end{abstract}

%% file: sections/1_introduction.tex
\section{Introduction}




Self-supervised learning (SSL) of visual representations~\cite{wu2018unsupervised,Ye_2019_end2end,oord2018cpcv1,tian2019cmc,he2019moco,chen2020simple,chen2020big,chen2020improved} has become a competitive alternative to supervised learning, without requiring manually annotated labels. 
A key component of SSL from images is contrastive learning, a learning objective that pulls different data augmentations from the same instances (known as query and key) to be closer to each other and pushes data augmentations from different instances away. However, not all of the commonly used augmentations in images reflect the visual variability that we see in the real world. 
\input{figures/open}

In contrast, videos provide a natural source of data augmentation, with objects undergoing 
deformations and occlusions, along with changes in viewpoints and illumination as shown in Figure \ref{fig:video_property}. As a result, recent work has tackled SSL from videos to seek more natural augmentations and meaningful semantics~\cite{pathak2017learning,owens2016ambient,misra2016shuffle,wang2017transitive,lee2017unsupervised,buchler2018improving,wei2018learning,jing2018self,sayed2018cross,purushwalkam2020demystifying,jabri2020walk}. A common approach \cite{gordon2020watching,chen2021rspnet} is to randomly sample nearby clips in videos as query and key as a natural way of data augmentation that represents the same instance since frames that are close in time are likely to share similar content. However, this sampling strategy for augmentation suffers from a few problems, as shown in Figure \ref{fig:missing_object} and \ref{fig:different_crop}. First, when sampling instances from a longer span of the video, the content might change substantially, resulting in samples containing totally different semantic concepts. This sampling strategy results in an imperfect supervisory signal that does not encourage semantic understanding.   Second, when sampling clips from the same video, the background in the two clips are often quite similar, which allows the model to cheat by looking at the background for minimizing contrastive loss~\cite{wang2021removing} as shown in Figure \ref{fig:same_scene}. This sampling strategy leads to models learning spurious background correlations and context, which could make them less transferable and potentially biased~\cite{choi2019can}.

To alleviate these problems, we propose \fullapp{} (\app{}). \app{} consists of an intelligent method to select query and key clips, which utilizes unsupervised tracking for videos. Using this freely available form of supervision, we design a temporal constraint for selecting clips that ensures that the query and the key contain the same object. In addition, using tracking information on the spatial extent of the object, we design spatial constraints to mask the background. Taken together, these spatial-temporal constraints result in better supervisory signals for contrastive learning from videos. After selecting more informative query and key clips, we train the model to learn to localize specific regions in query and key that represent the same concepts using a Grad-CAM~\cite{selvaraju2017gradcam}-based attention loss.  
We pretrained a momentum contrastive encoder (MoCo)~\cite{he2019moco} with \app{} on Image and Video-based SSL backbones using VGG-Sound and Kinetics-400 datasets. Evaluation on video downstream tasks, including action recognition, video retrieval shows that \app{}-trained models learn more accurate visual representations. In particular, we obtain state-of-the-art performance on video action classification. Due to its ability to localize objects, \app{}-trained models can perform unsupervised tracking across arbitrary lengths of videos, as shown by our experiments on 
the DAVIS challenge~\cite{perazzi2016davis}. 
Additional experiments on image and video datasets with background changes show that models trained with \app{} are less dependent on background correlations and are more robust to background changes in visual classification. We also showed the various invariances (occlusion, viewpoint) captured by our model.

In sum, our work shows that existing methods for contrastive SSL from videos do not efficiently use temporal transformations of objects. By designing a better clip sampling strategy and a loss that encourages object localization, we are able to learn more accurate visual representations from the video that are robust to background changes. 


%% file: figures/open.tex
    


\begin{figure}[!t]
    \begin{subfigure}[t]{\columnwidth}
    \centering
    \includegraphics[width=\textwidth]{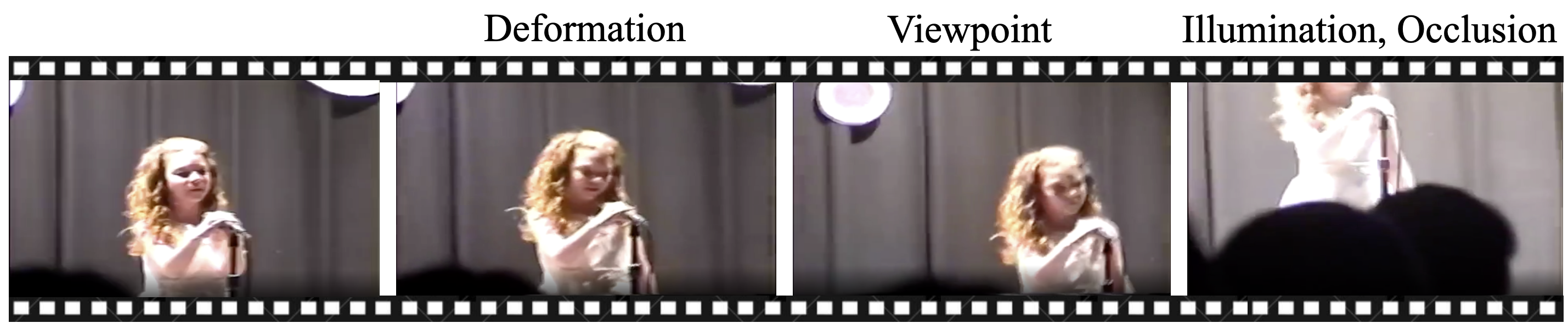}
    \caption{Temporal transformations in videos provide a natural source of data augmentation, making them attractive for self-supervised learning (SSL).}
    \vspace{+10pt}
    \label{fig:video_property}
    \end{subfigure}
    \vspace{+10pt}
    \begin{subfigure}[t]{\columnwidth}
    \includegraphics[width=\textwidth]{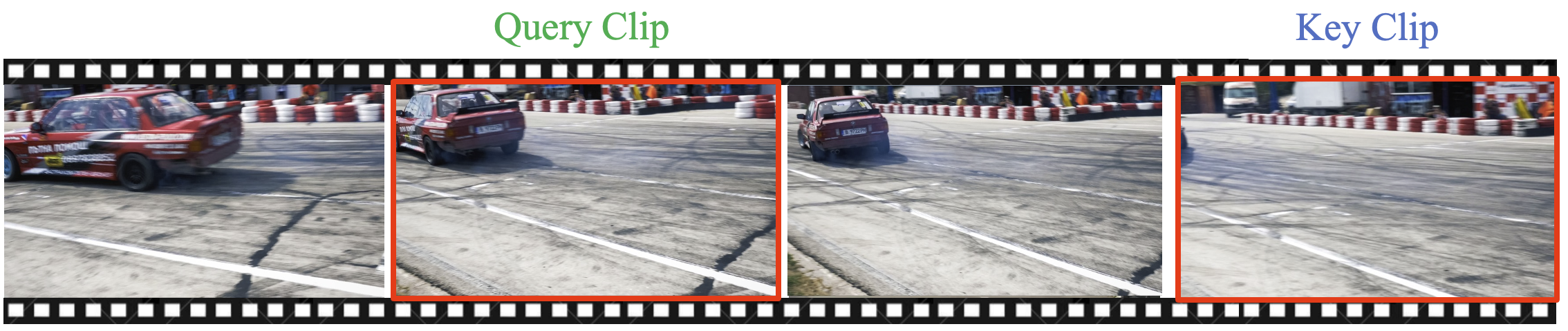}
    \caption{Randomly selected query and key clips in contrastive video SSL may lead to missing objects.}
    \label{fig:missing_object}
    \end{subfigure}
    \vspace{+10pt}
    \begin{subfigure}[t]{\columnwidth}
    \includegraphics[width=\textwidth]{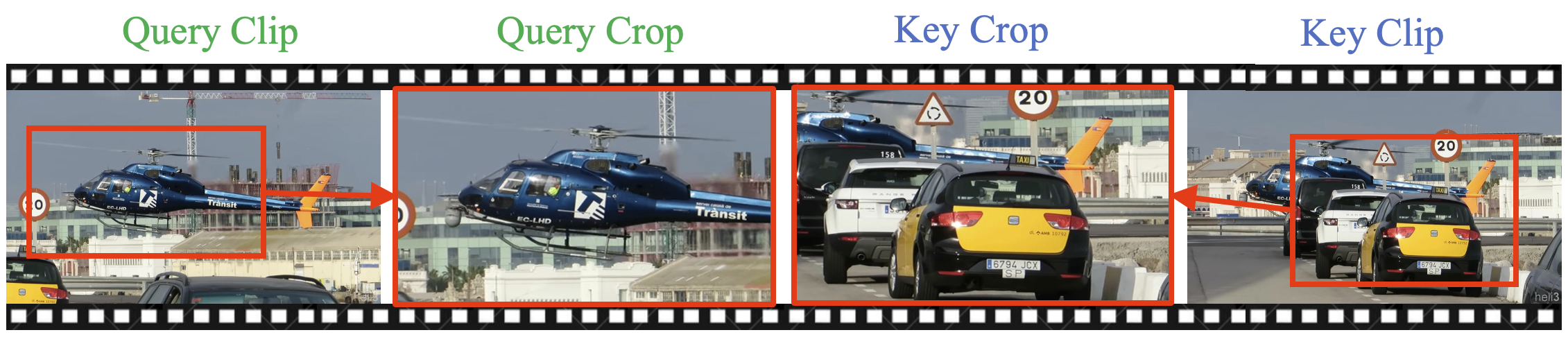}
    \caption{Query and key clips may also contain different visual concepts altogether.}
    \label{fig:different_crop}
    \end{subfigure}
    \vspace{+10pt}
    \begin{subfigure}[t]{\columnwidth}
    \includegraphics[width=\textwidth]{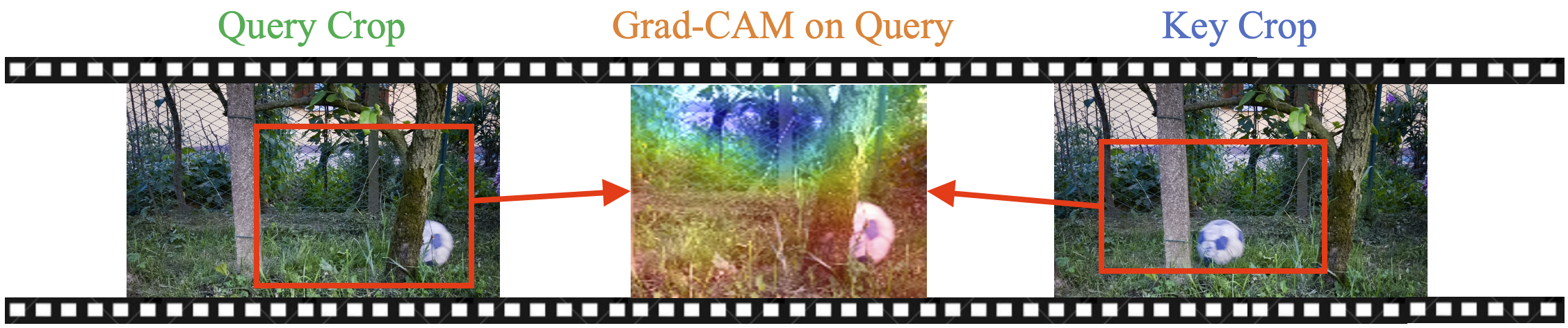}
    \caption{Since many videos contain a fixed background, SSL models can cheat by focusing on the background.}
    \label{fig:same_scene}
    \end{subfigure}
    
    \caption{Current methods for contrastive video self-supervised learning receive an imperfect supervisory signal and can rely on background correlations when learning representations. We propose a new approach by video tracking and Grad-CAM supervision to tackle these problems. 
    }
	\label{fig:problems_with_ssl}
	\vspace{-10pt}
\end{figure}

%% file: sections/2_related_work.tex
\section{Related Work}

\paragraph{Self-supervised representation learning (SSL).}
Contrastive SSL approaches learn image representations ~\cite{bachman2019learning,henaff2019data,tian2019cmc,zhuang2019local,chen2020simple,misra2020self,benaim2020speednet} by forming positive and negative pairs, and
maximizing the similarity of positive pairs as compared to negative pairs. 
Positive pairs are generated from a single image instance through artificial data augmentations such as random cropping, resizing, color distortion, and Gaussian blur~\cite{chen2020simple}. 
Going beyond learning representations from images, different frames of videos provide natural viewpoint changes and temporal information which can help learn better representations in a self-supervised manner ~\cite{Agrawal2015,Wang_UnsupICCV2015,pathak2017learning,wang2017transitive,vondrick2018tracking,qian2021spatiotemporal,ranasinghe2022self,hu2021contrast,qing2022learning}. 
Saliently, contrastive learning-based methods~\cite{gordon2020watching,purushwalkam2020demystifying,jabri2020walk,xu2021rethinking,dwibedi2019temporal} that sample positive pairs from the same video have shown that view-point invariant representations can be learnt from videos. \bccomment{Unlike previous methods~\cite{Wang_UnsupICCV2015,purushwalkam2020demystifying} that sample positive pairs from unsupervised proposals with bounding boxes, we introduce an approach for sampling pairs based on spatial and temporal constraints obtained using unsupervised saliency maps,} coupled with Grad-CAM supervision~\cite{selvaraju2017gradcam} to learn better grounded representations. 

\paragraph{Grounded Representation Learning.}
Our work is also related to recent work on learning better grounded representations.
Henaff \etal \cite{henaff2021efficient} introduced DetCon, a self-supervised objective which tasks representations with identifying object-level features across different image augmentations. 
Mo \etal \cite{mo2021object} introduced a technique to mix backgrounds of different images during contrastive pretraining and showed that it leads to models learning reduced contextual and background biases. 
Xie \etal \cite{xie2021unsupervised} propose an object-level pretraining approach for learning from complex scenes. CAST~\cite{selvaraju2021casting} learns visually grounded representations through saliency supervision. 
FAME~\cite{ding2022motion} extracts moving foreground by frame difference and color statistics to alleviate background bias.

%% file: sections/3_method.tex
\section{Method}
We propose \fullapp{} (\app{}) to learn visual representations from videos by utilizing unsupervised object tracking. 
First, we will review the standard contrastive based video representation learning framework and then discuss our approach.

\input{figures/pipeline}

\subsection{Background}
When performing contrastive learning on videos, the positive pairs are clips from the same video selected from different times, while the negative pairs are formed with clips taken from other videos. 
In this work, we build our approach on top of the Momentum Contrast (MoCo)~\cite{he2019moco} model, which uses the InfoNCE~\cite{oord2018cpcv1} objective and stores the negative samples in a dynamic memory bank with a moving average encoder.
Formally,  given a video $V$, we learn feature representations for query $q$ and key $k$ sampled from the same video. 
The goal is to pull the feature distance of the positive pairs $q$ and $k$ to be closer and push the features of query $q$ away from a negative set of features from other videos $N = \{n_1, n_2, ..., n_m\}$.
The MoCo loss is:
\begin{equation}
\small
    \mathcal{L}_{\text{MoCo}} =-\log \frac{\exp \left(q \cdot	 k ) / \tau\right)}{\sum_{n \in \{N, k\}} \exp \left(q \cdot n ) / \tau\right)},
    \label{eq:intra-video}
\end{equation}
where $\tau$ is the temperature constant. 

In the video model, in addition to the MoCo loss, we also use the \textbf{relative speed prediction task} which has been found to be beneficial to understand the relative speed between the video segments proposed in \textbf{RSPNet}~\cite{chen2021rspnet}. We sample three video segments: two segments with the same speed and another with a different speed. The goal is to pull the feature distance for segments with the same speed closer together while pushing the features for the segment with different speed away. A triplet loss~\cite{tripletloss} is applied:
\begin{equation}
\small
	\label{eq:ranking}
	\mL_{Speed} = { {\rm max}(0,  \gamma - (pair^{+} - pair^{-})) },
\end{equation}
where the distance of positive pairs  $pair^{+}$ should be larger than the negative pairs $pair^{-} $ by a margin $\gamma >0$.



\subsection{Unsupervised tracking in videos}
\label{sec:constrained_cropping}
In order to select query and key clips from the same video that contain the same visual concepts, we propose to use unsupervised object tracking to guide clip selection. 
To acquire unsupervised tracking information from the video we first use Deep-USPS~\cite{nguyen2019deepusps}, an unsupervised saliency prediction algorithm, to obtain a saliency map for the initial frame in the video. 
We use this saliency map as the target object for tracking and apply SORT \cite{bewley2016simple}, a tracking algorithm which checks the IoU constraint across continuous frame masks to track the target object through the video. 
Formally, given an input video $V$ with height $h$, width $w$ and temporal length $t$, we acquire the video object segmentation map $M \in \{0, 1\}^{h \times w \times t}$, where $M_{ijk} = 1$ indicates pixel $(i, j, k)$ is salient, and area of salient region in time $t$ is $A_M^t = \sum_{i,j} M_{i,j}$. 
The saliency map is a binary mask. Since a large majority of the web videos (and as a result, videos in vision datasets) are centered on a single object, we only utilize one (the largest) salient region in the video for tracking and do not consider multiple objects in this work. 

\noindent\textbf{Spatial-temporal cropping based on video tracking:} 
Once we obtain the tracking tube for the video, we constrain our random sampling to video segments covered by the tracking tube as shown in left half of Figure \ref{fig:pipeline}, where $A_M^t \neq 0$. This ensures that our sampled query and key clips contain meaningful instances of the same object in the video. In addition, we set a spatial constraint (\Cref{fig:pipeline}): the random crop for the query or key should have at least $\mu \in [0, 1)$ IoU with the tracking mask. This spatial constraint tries to ensure that the query and key contain the same object for contrastive pretraining. 
We acquire two 3D masks for the video segment $M_q$ and $M_k$, which represent the mask of the \textit{query} and \textit{key} containing salient regions. 



\subsection{\fullapp{} (\app{})}
\app{} aims to encourage the model to learn to localize specific regions within the query and key that represent the same concept. 
We first determine the regions that the network relies on when matching the object regions in the key, $x^k$ with that of the query, $x^q$. To obtain the object regions in key, we mask the key with the video segmentation mask, $M_k$, as a filter to get the key foreground, $x^{k_m}  = x^k * M_k$. 
To understand the importance placed by the network on specific crop regions when contrastively matching their representations, 
we compute Grad-CAM~\cite{selvaraju2017gradcam} in a contrastively-trained fashion. 
We do this by first forward propagating the key foreground, $x^{k_m}$, and the query, $x^q$, through the respective encoders to get $k^m$ and $q$. To get the regions that would help maximizing their similarity, we take their dot-product and compute the gradients \textit{wrt} the last convolution layer activations of the query encoder, $f_q$, as follows:

\begin{equation}\label{eq:alpha}
\small
    \alpha{}_{q} =
    \overbrace{
        \sum_{i,j}
    }^{\text{global pooling}} \mkern-65mu
    \hspace{10pt}
    \underbrace{         
        \frac{\partial q \cdot k^{m}}{\partial A_{conv5}^{f_q}}
    }_{\text{gradients via backprop}}
\end{equation}
where the $\alpha_q$ represents the last convolutional layer neurons' importance for maximizing the similarity of the query and the key foreground representations. 
Through a weighted combination of $\alpha_q$ with the last convolutional layer activations $A_{conv5}^{f_q}$ and clipping them at zero, we can get Grad-CAM maps, $\text{ \cal G}_q$. 
\begin{equation} \label{eq:gcam}
\small
    \text{ \cal G}_q = ReLU \underbrace{\left(\sum_{n} \alpha{}_{\text{q}} A_{conv5}^{f_{\text{q}}}\right)}_{\text{linear combination}}.
\end{equation}
Higher values in $\text{ \cal G}_q$ represents the regions the network relies on when mapping query to key foreground.

We would ideally want the network to only rely on the tracked object regions in the query that are highlighted in the key foreground. 
Therefore, we apply a cosine-distance based attention loss to encourage the Grad-CAM heatmap $\text{ \cal G}_q$ to be close to tracked object mask in the query segment $M_q$. 
This enforces the model to learn similar representations for the object irrespective of the viewpoint and transformation changes that might be present in the clips when the frames are temporally far away.
\bccomment{We interpolate $M_q$ to the same spatial and temporal dimension as $\text{ \cal G}_q$ to acquire the pseudo segmentation ground-truth, $\hat{M_q}$ as the supervision for the Grad-CAM heatmap.}
The Attention loss is defined as:
\vspace{10pt}

\label{sec:previts_loss}
\begin{equation}
\small 
{\cal L_{\text{att}}} =  1 - \frac{ G_{q} \cdot \hat{M_{\text{q}}}} {\lVert G_{q} \rVert~\lVert \hat{M_{\text{q}}}\rVert}.
\end{equation}


Our full model is trained to minimize the sum of the losses described above.
\begin{align}
\small
    {\cal L_{\text{Total}}} = \cal L_{\text{MoCo}}+\cal L_{\text{Speed}} +\lambda \cal L_{\text{Att}}.
    \label{eq:loss}
\end{align}



%% file: figures/pipeline.tex
\begin{figure*}[t]
    \centering
    \includegraphics[width=\textwidth]{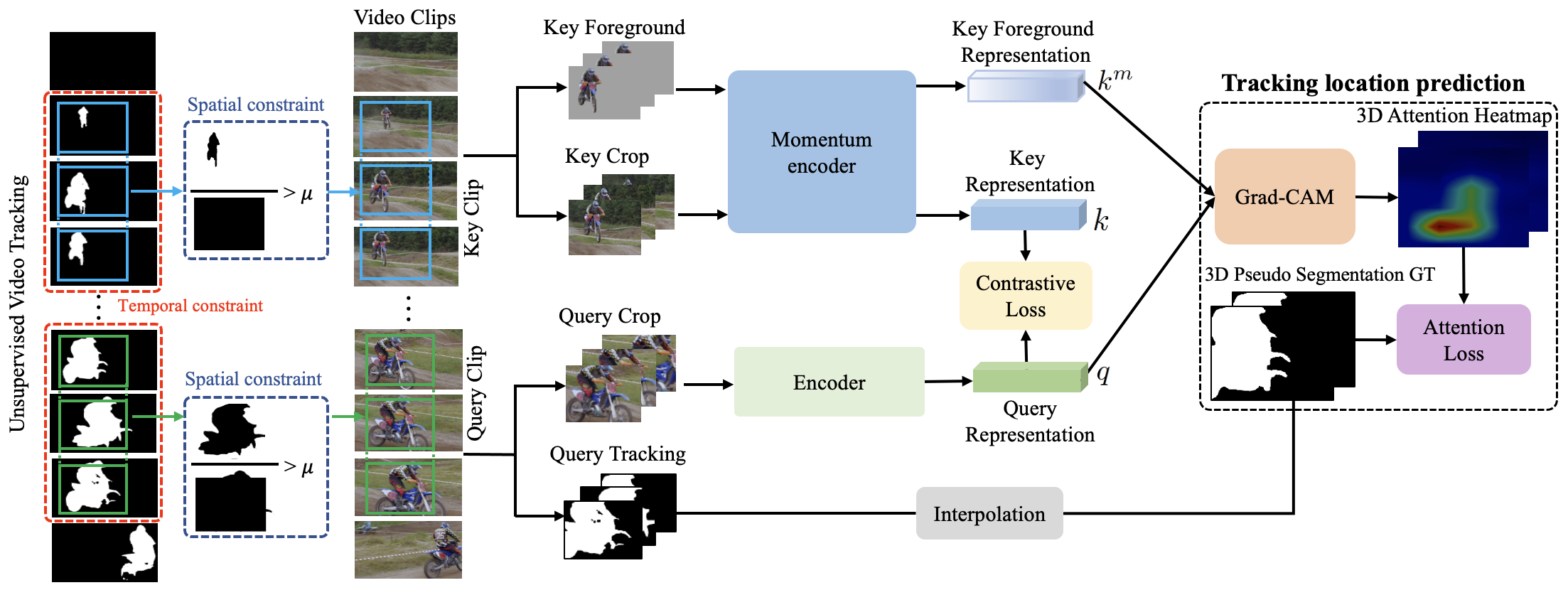}
    
    \caption{\textbf{Pretraining with Video Tracking Supervision (PreViTS):} Given an input video, we  perform unsupervised tracking and apply temporal constraints to extract continuous frames that contain the tracked object region. We then apply IoU based spatial constraints to sample query and key video clips along with their masks. The encoder representations for the query and key are aligned through a contrastive loss. We then mask the key and use Grad-CAM to localize the regions in the query that maximize the (key foreground, query) similarity. We then supervise Grad-CAM with the tracked query mask using a cosine distance loss to encourage models to rely on appropriate salient object regions during contrastive pretraining.}
    \label{fig:pipeline}
\end{figure*}

%% file: sections/4_experiments.tex
\section{Experiments}
We aim to show that training video self-supervised models with \app{} leads to better representations that obtain improved transfer learning performance with reduced dependence on background signal and context. We validate this by pretraining representations on two datasets and transferring them to various video and tracking tasks.

\subsection{Implementation details}
We pretrain our models on two datasets independently, both consist of 10 second-long videos at 25 FPS: (1) The  \textbf{VGG-Sound} \cite{chen2020vggsound} dataset contains
$~$200k videos collected from YouTube. VGG-Sound was collected with the objective of creating an audio-visual dataset with diverse sounds and contains 300 classes as defined by audio labels. 
\bccomment{Unlike previous video SSL methods that test on video downstream tasks, we also learn object concepts from videos for image understanding. So we chose VGG-Sound, which contains a wider variety of object classes and higher object-centricity as compared to action classification datasets common in the video understanding literature. Also, a large majority of VGG-Sound videos only contain a single foreground object, as we found by using supervised segmentation, which is consistent with our single object assumption in the learning phase.} (2) The \textbf{Kinetics-400} dataset~\cite{carreira2017kinetics} is a widely-used dataset, which enables us to compare \app{}'s performance to prior methods. It consists of around 240k training videos with 400 human action classes.
We will release the code for replicating our work. More details and image recognition experiments can be found in the supplement.

\input{tables/video_ex}

\subsection{Video tasks}
\label{sec:downstream_eval_vid}
\noindent\textbf{Action recognition:} To evaluate the performance of \app{}-trained models on video classification tasks, we perform action recognition on the UCF-101 dataset \cite{soomro2012ucf101}. Following Xu \etal~\cite{xu2019self}, in all experiments, we finetune our pretrained model on labeled videos with 50 epochs using a learning rate of 0.05. We drop the projection head and replace it with a randomly initialized fully-connected layer. We report top-1 accuracy on the UCF-101 dataset when pretraining with \app{} on VGG-Sound and Kinetics-400 datasets in Table~\ref{tab:main_results_video}. Training with \app{} obtains a substantial improvement over RSPNet on both pretraining datasets. 
Notably, the model pretrained on Kinetics-400 had better performance with RSPNet and a larger absolute improvement with RSPNet + \app{} (4.2\% versus 2.5\%), over VGG-Sound. We speculate that since human actions are better represented in Kinetics-400, the representation learnt using these videos transfers better to UCF-101, and also benefits more from training with \app{}. 
Finally, we compare the performance of RSPNet + \app{} pretrained with Kinetics-400 with other state-of-the-art video SSL methods~\cite{chen2021rspnet} in Table \ref{tab:main_results_sota}. With the same architecture, computational budget, epoch, batch size, and pretraining data for a fair comparison, our approach outperforms prior work and obtains state-of-the-art performance. 
\input{tables/video_sota}

\input{tables/retrieval}

\noindent\textbf{Video retrieval: } \bccomment{We also evaluate our video retrieval task on the UCF-101 dataset.
Given a video as a query, we search the most relevant video by cosine distance using the nearest neighbor search. Following \cite{chen2021rspnet}, we evaluate our method on the split 1 of UCF101 dataset and apply the top-$k$ accuracies ($k$=1, 5, 10, 20, 50) as evaluation metrics.
As shown in Table~\ref{tab:retri_acc}, our model outperforms the other baselines by a large margin, showing the effectiveness of the proposed training process.}
%
\input{figures/video_back}

\input{tables/image_bg}

\input{tables/invariance_video}

\subsection{Backgrounds challenge} 
We expect feature representations obtained using \app{} to be less dependent on object backgrounds and context. To quantify this, we utilize the ``backgrounds challenge'' \cite{xiao2020noise} on both image and video classification tasks as shown in \Cref{tab:backgrounds_results}. 

\paragraph{Backgrounds Challenge.}
First, we evaluate our model on the original Backgrounds Challenge~\cite{xiao2020noise}, which was designed to test a model's robustness to various background changes. It contains 9 ImageNet classes with 450 images for each class. We evaluate our model along with the baseline model pretrained on VGG-Sound and train a linear layer with ImageNet-1K. Results show that pretraining with \app{} achieves significant improvement on all tasks defined in the Backgrounds Challenge. Examples of different settings can be found in Figure \ref{fig:video_back}. In the Only-FG setting, where the background is set to black, \app{} obtains an absolute improvement of $12.1\%$, showing that it is less dependent on background information. When backgrounds are swapped (Mixed-Same, Mixed-Rand, Mixed-Next), \app{} obtains an absolute improvement of $3.6-4.2\%$, indicating that representations learnt with \app{} reduce the reliance on background correlations. There is a slight increase in performance in the No-FG setting, likely due to the model learning contour information from videos. However, in settings where no information from the foreground is provided (Only-BG-B, and Only-BG-T), \app{} obtains lower accuracy than baseline, which reinforces that it is less dependent on the background signal.

\paragraph{Video Backgrounds Challenge (JHMDB).}
Taking inspiration from the Backgrounds Challenge dataset, we construct a new Video Backgrounds Challenge to test background-robustness on videos. We use the JHMDB dataset \cite{jhuang2013jhmdb}---consisting of 21 HMDB \cite{hmdb51} action recognition classes with 50 videos per class---for which the ground truth foreground mask is available. We follow Xiao et al.~\cite{xiao2020noise} to construct 8 foreground-background combinations (\reffig{fig:video_back}) for JHMBD. We evaluate performance using a model trained on Kinetics-400 and finetuned on UCF-101 \bccomment{and JHMDB}. Models trained with \app{} outperform the baseline model  (RSPNet) in all settings. Similar to the trends on Backgrounds Challenge, \app{} obtains significant improvement in settings where the background is set to black or is replaced by background from another video. 
In settings where the foreground is removed, we find the accuracy drop to be higher for \app{} compared to baseline (22.1 vs. 21.6).
Video representation learning models have been shown to suffer from over-reliance on background information, called representation bias~\cite{li2018resound} or scene bias~\cite{choi2019can}. Training with \app{} can help mitigate this bias.  


\subsection{Invariances captured by \app{}.~}
{We expect representations learnt by \app{} to have better invariance to various transformations (occlusion, viewpoint, illumination, instance), due to more effective use of object instance information during contrastive learning.
Following \cite{purushwalkam2020demystifying}, we  measured the representation's  invariances when predicting classes using the top-k Representation Invariance Score (RIS). We selected top-10/25 neurons from encoder with similar activation behavior between transformations and computed its mean score. \app{} is significantly more invariant to transformations than other baselines (\Cref{tab:invariance}).}

\subsection{Video tracking evaluation}
To demonstrate grounding and tracking ability, we evaluate \app{} on single object video tracking~\cite{perazzi2016davis} in Grad-CAM attention fashion. 
In the original video tracking task, the input is the first frame of the video along with the foreground segmentation mask. The goal is to predict the pixel-level mask of the foreground in the later video frames. In our setting (\reffig{fig:pipeline}), we feed the first frame and its segmentation to acquire the key foreground. Then, we feed the later frames as queries and compute the Grad-CAM attention heatmap to localize the corresponding region in the later frames. Since the attention heatmap resolution is $7\times 7$, we cannot perform pixel-level prediction. 
Our evaluation metrics follow \cite{perazzi2016davis}. We compute: Region similarity ($\cal J$), which represents the IoU between the predicted foreground mask and GT foreground mask; Mean ($\cal M$) is the average value of $\cal J$; Recall ($\cal O$) evaluates the fraction of sequences scoring higher than a threshold; Decay ($\cal D$) evaluates the averaged performance drop over time, e.g., $\cal J_{\text{t=4}} - \cal J _{\text{t=1}}$. \app{} outperforms the baseline MoCo by a significant margin (\Cref{tab:davis}), which demonstrates our model's ability to localize objects in dynamic videos. \reffig{fig:davis} shows how \app{} is able to localize objects while the baseline fails when the object appears in a novel viewpoint (\reffig{fig:davis}(d)).

\input{tables/davis}

\input{figures/davis}

\input{tables/ablation}

\input{figures/visualization}
\input{figures/occlusion}
\subsection{Ablations and Analysis}
We conduct an ablation study on the effect of our design decisions. We evaluate UCF-101 trained on K400 for 50 epochs following~\cite{patrick2021gdt}. We also tested the image dataset \voc{} object detection \cite{everingham2009voc} trained on the VGG-Sound for 200 epochs.
More details of the image model can be found in the supplement.

\noindent \textbf{Temporal distance constraint between positive pairs:}
We investigate the effect of different temporal sampling strategies in \Cref{subtab:ablations1}. We define $\delta$ to be the temporal distance between the query and key segment.  
\textbf{$\delta =0$} uses the same sample segment for query and key with image augmentation.
\textbf{Constant $\delta$} samples query and key in a fixed length of 1.7 sec, which ends up as an easier task for the model and does not generalize to the downstream task.
\textbf{Varying $\delta$} does not constrain the distance between the clips, \bccomment{which refers to random sampling query and key without additional temporal distance constraint}. We find this setting to perform the best as it enables the network to localize regions across the clips irrespective of their temporal distance. 


\noindent \textbf{Effect of area threshold $\mu$:}
We apply spatial constraint when sampling our positive pairs where the crop covers at least $\mu$ IoU of the tracking object area. Here, we investigate the different values of $\mu$ in the range 0 to 0.9. Results in \Cref{subtab:ablations2} demonstrate that adding spatial constraint helps the model focus on meaningful objects in the video. \bccomment{We also find the performance result is not sensitive to value of $\mu$, demonstrating consistent improvement of our method.}

\noindent \textbf{Effect of loss weight $\lambda$:}
We test different loss weights $\lambda$ to balance between the two losses.
Results from \Cref{subtab:ablations2} show that non-zero values of $\lambda$ outperform $\lambda$ = 0.0, indicating that attention loss is important in \app{}. Higher $\lambda$ improves performance up to a point---performance improves with $\lambda$= 2.0, 3.0, and slightly degrades with $\lambda$= 4.0. We find $\lambda=3.0$ to be optimal.


\noindent \textbf{Robustness to the quality of tracking mask: }
To understand the effect of the quality of tracking supervision, 
\bccomment{
we experimented with a lower quality tracking mask by replacing segmentation masks with a bounding box, which is less accurate in terms of the shape of the object (\Cref{subtab:ablations4}). The model obtained significant gain on \voc{} over baseline (\textcolor{ForestGreen}{\textbf{+3.6}}) and (\textcolor{Gray}{\textbf{-1.1}}) compared to our best model.}
Our model trained with unsupervised tracking mask still achieves comparable performance with the model using the supervised segmentation, which demonstrates its robustness to noises generated from unsupervised tracking.



\noindent \textbf{Visual grounding and localization:} We also visualize the grounding and localization ability of \app{}-trained models finetuned on UCF-101 using Grad-CAM. Our model has a better grounding ability as compared to the baseline and focuses on foreground objects instead of background scenes (\reffig{fig:ucf}). In \reffig{fig:occlusion}, we provide a query with two different segmentation corresponding to the different foreground objects. We feed the query and the key foreground into the \app{}-trained model to compute the Grad-CAM attention heatmaps. Given the different key foreground, our model can localize the man and ball, respectively. At the same time, the attention heat map in the baseline is more spread out and cannot generate discriminative attention of the two objects. Even though \app{} hasn't seen multi-object masks during pretraining, it is still able to localize multiple concepts discriminatively. 

%% file: tables/video_ex.tex
\begin{table}[h]
    \newcommand{\apbbox}[1]{AP$^\text{bbox}_\text{#1}$}
    \newcommand{\apmask}[1]{AP$^\text{mask}_\text{#1}$}
    \newcolumntype{Y}{>{\raggedright\arraybackslash}X}
    \newcolumntype{Z}{>{\centering\arraybackslash}X}

    \centering
    \footnotesize
    \setlength\tabcolsep{1pt}
    \renewcommand{\arraystretch}{1.2}

    \begin{tabular}[]{c l c c c c}
    \toprule
    ~
    & \multicolumn{1}{l}{\bf \multirow[b]{1}{*}{Method}}
    
    & ~~~
    & 
    \multicolumn{1}{l}{\bf \multirow[b]{1}{*}{Dataset}}
    & ~~~
    &
    \multicolumn{1}{c}{\bf UCF-101}
     \\

    \midrule
    
    & RSPNet & & VGG Sound && 86.4 \\

      & \hspace{5pt} + Tracking Constrained Sampling && VGG Sound  && 87.5\Rise{1.1} \\
    
     & \hspace{5pt} + \app{} && VGG Sound  && 88.9\Rise{2.5} \\
    \midrule
      & RSPNet & & K400 && 87.6 \\

      & \hspace{5pt} +  Tracking Constrained Sampling && K400  && 89.1\Rise{1.5} \\
    
      & \hspace{5pt} + \app{} && K400  && 91.8\Rise{4.2} \\
                
    \bottomrule
    \end{tabular}
    \caption{\textbf{Video Action Classification:} \app{} obtains significant performance gains on the commonly-evaluated downstream task of UCF-101 action recognition. Tracking Constrained Sampling refers to our unsupervised tracking-based spatial-temporal sampling strategy. 
    } 
    \label{tab:main_results_video}
\end{table}

%% file: tables/video_sota.tex
\begin{table}[t]

    \centering
    \footnotesize
    \begin{tabular}[]{@{} l c c c c @{}}
    \toprule
     
      \textbf{Method}
     & 
     \textbf{Input size}
     & 
     \textbf{Params}
     & 
     
     \textbf{Backbone}
     & 
     
     \textbf{UCF-101}
      \\
       \midrule
       RSPNet \cite{chen2021rspnet} & 112 $\times$ 112 & 33.4M &  C3D & 76.7 \\
       CACL \cite{guo2022cross} & 112 $\times$ 112 & 33.4M &  C3D & 77.5 \\
       \textbf{\app{}} & 112 $\times$ 112 & 33.4M &  C3D & \textbf{78.7} \\
       \midrule
       
       Pace \cite{wang2020self} & 112 $\times$ 112 & 14.4M&  R(2+1)D & 77.1 \\
       STS \cite{wang2021self} & 112 $\times$ 112 & 14.4M&  R(2+1)D & 77.8 \\
       VideoMoCo \cite{pan2021videomoco} & 112 $\times$ 112 & 14.4M&  R(2+1)D & 78.7 \\
       RSPNet \cite{chen2021rspnet} & 112 $\times$ 112 & 14.4M&  R(2+1)D & 81.1 \\
       \textbf{\app{}} & 112 $\times$ 112 & 14.4M &  R(2+1)D & \textbf{81.9} \\
       \midrule
       
      SpeedNet \cite{benaim2020speednet} & 224 $\times$ 224 & 9.6M &  S3D-g  & 81.1 \\

      CoCLR \cite{Han20} & 224 $\times$ 224 & 9.6M &  S3D-g  & 87.9 \\
      STS \cite{wang2021self} & 224 $\times$ 224 & 9.6M&  S3D-g & 89.0 \\
      RSPNet \cite{chen2021rspnet} & 224 $\times$ 224 & 9.6M&  S3D-g & 89.6 \\
      ASCNet \cite{Huang_2021_ICCV} & 224 $\times$ 224 & 9.6M&  S3D-g & 90.8 \\
      \textbf{\app{}} & 224 $\times$ 224 & 9.6M &  S3D-g & \textbf{91.8} \\
    \bottomrule
    \end{tabular}
    \vspace{+5pt}
    \caption{\textbf{Comparison to prior work on UCF-101 performance:} Our best-model trained with \app{} outperforms all existing methods for video self-supervised learning on UCF-101 downstream performance, when using comparable training resources. 
    } 
    \label{tab:main_results_sota}
\end{table}

%% file: tables/retrieval.tex
\begin{table}[t]
	\centering
	\scriptsize
	\begin{tabular}{lccccc}
		\hline
		\multirow{2}[0]{*}{Method}  &
		\multicolumn{5}{c}{Top-$k$} \\
		\cline{2-6}
		&  $k=1$    & $k=5$    & $k=10$   & $k=20$   & $k=50$     \\ 
		\hline
		Pace~\cite{wang2020self}                   & 31.9  & 49.7  & 59.2 & 68.9  & 80.2   \\
		\multirow{1}{*}{RSPNet \cite{chen2021rspnet}}    
		              & 36.0           & 56.7          & 66.5          & 76.3           & 87.7   \\
		\multirow{1}{*}{STS \cite{wang2021self}}    
		               & 39.1          &  59.2         & 68.8         & 77.6          & 86.4   \\
		\multirow{1}{*}{CACL \cite{guo2022cross}}    
		               & 43.2          &  61.1         & 69.9         & 78.2          & 88.2   \\
		\multirow{1}{*}{TCLR \cite{dave2021tclr}}    
		              & 48.6           & 67.6         & 75.5          & 82.5         & -   \\
		\multirow{1}{*}{\textbf{\app{}}}    
		             & \textbf{53.4}           & \textbf{69.4 }         & \textbf{77.8}         & \textbf{85.5}           & \textbf{93.0}   \\
		\hline
	\end{tabular}
	\vspace{+0.1cm}
	\caption{\textbf{Video retrieval} results on UCF101. Our model outperforms other baselines using the same architecture C3D backbone.\label{tab:retri_acc}}
\end{table}


%% file: figures/video_back.tex
\begin{figure}[t]
    \centering
    \includegraphics[width=.48\textwidth]{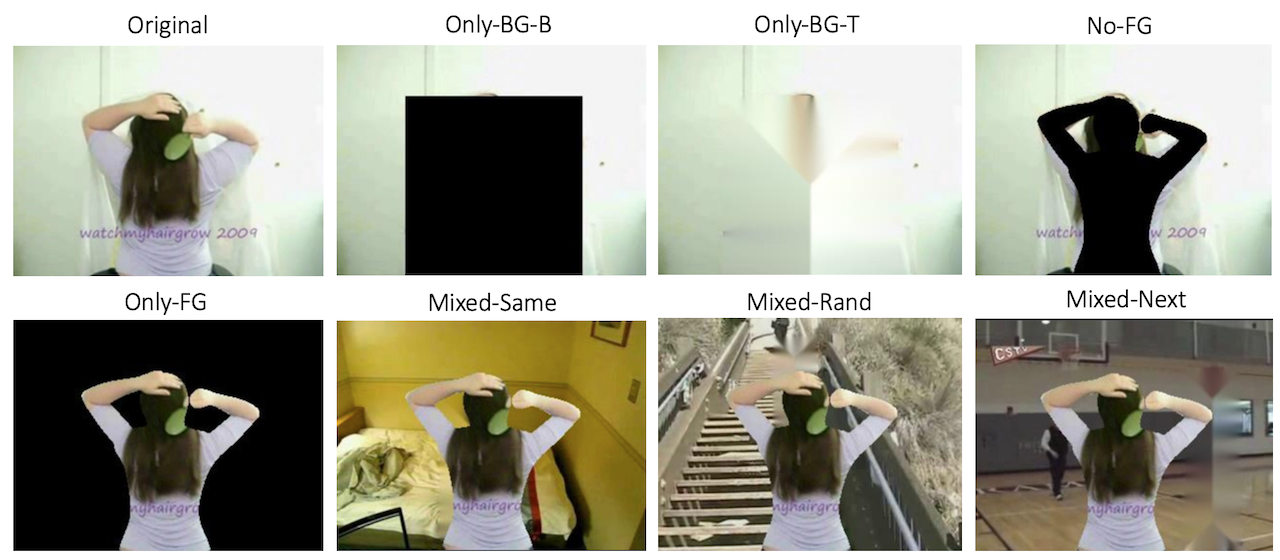}
    \caption{\textbf{Video Background Challenge:} We evaluate \app{} by introducing a Video Backgrounds Challenge to evaluate background-robustness of video models. FG = foreground, BG = background. Foreground-background combinations include: Only-BG-B (FG:  Black, BG: Unmodified), Only-BG-T (FG: Tiled background, BG: Unmodified), Mixed-Same (FG: Unmodified,  BG: Random BG of the same class), Mixed-Rand (FG: Unmodified, BG: Random BG of a random class), and Mixed-Next (FG: Unmodified, BG: Random BG of the next class.)}
    \label{fig:video_back}
\end{figure}

%% file: tables/image_bg.tex
\begin{table*}[t]

    \newcolumntype{Y}{>{\raggedright\arraybackslash}X}
    \newcolumntype{Z}{>{\centering\arraybackslash}X}

    \centering
    \footnotesize
    \renewcommand{\arraystretch}{1.2}

    \begin{tabularx}{\textwidth}{ p{0.19\textwidth} c llllllll}
    \toprule
    &  & \multicolumn{8}{c}{\bf Backgrounds Challenge~\cite{xiao2020noise}}\\
    \midrule
     \multicolumn{1}{l}{\textbf{VGG Sound}} &  ~& Original & \makecell{Mixed-Same}  & \makecell{Mixed-Rand}  & \makecell{Mixed-Next} &   \makecell{Only-FG}  &
     \makecell{No-FG} & \makecell{Only-BG-B}  &  \makecell{Only-BG-T}    \\
    \midrule
    MoCo &~& 77.9 & 53.3 & 37.8 & 33.8 & 40.9 & 24.6  &  9.7 & 13.5  \\
    \app{}  &~&      81.0\Rise{3.1}   &  56.9\Rise{3.6}     & 42.0\Rise{4.2}      & 38.0\Rise{4.2} & 53.0\Rise{12.1} &  28.0\Rise{3.4}         &  8.8\Riseneg{0.9}    &  13.0\riseneg{0.5}   \\
    \midrule
    \multicolumn{1}{l}{\bf \multirow[b]{1}{*}{\makecell{K400 \\}}}
    & 
    & \multicolumn{8}{c}{\bf Video Backgrounds Challenge}\\
    \midrule
    RSPNet &~& 70.7 & 40.7 & 30.3 & 29.5 & 20.9 & 49.1  &  35.2 & 28.6  \\
    \app{}   &~&      74.0\Rise{3.3}   &  48.0\Rise{7.3}     & 35.9~\Rise{5.6}    & 32.7\Rise{3.2}    &  27.8\Rise{6.9}   & 51.9\Rise{2.8}   &  33.7\Riseneg{1.5}    &  28.3\riseneg{0.3}   \\
    \bottomrule
    \end{tabularx}
    \caption{\textbf{Robustness to background changes}. On image and video Backgrounds Challenge datasets, \app{} outperforms baselines where the foreground was included (columns 1-5), especially the Only-FG setting. Also, \app{}-trained models are less accurate when foreground information is entirely eliminated (columns 7, 8), showing their reduced reliance on background information. }
    \label{tab:backgrounds_results}
    \vspace{+10pt}
\end{table*}

%% file: tables/invariance_video.tex
\begin{table*}[h]
\centering
\resizebox{\linewidth}{!}{
\scriptsize
\begin{tabular}{lcccccccccccc}
\toprule
\multirow{2}{*}{\textbf{Method}} & \multicolumn{2}{c}{\textbf{Occlusion}} & \multicolumn{2}{c}{\textbf{Viewpoint}} & \multicolumn{2}{c}{\textbf{Illumination Dir.}} & \multicolumn{2}{c}{\textbf{Illumination Color}} & \multicolumn{2}{c}{\textbf{Instance}} & \multicolumn{2}{c}{\textbf{Instance+Viewpoint}} \\
 & Top-10 & Top-25 & Top-10 & Top-25 & Top-10 & Top-25 & Top-10 & Top-25 & Top-10 & Top-25 & Top-10 & Top-25 \\

\midrule
MOCO & 83.25 & 76.45 & 84.83 & 75.31 & 85.09 & 74.74 & 99.42 & 95.88 & 48.99 & 43.55 & 51.23 & 46.83 \\
Region Tracker \cite{purushwalkam2020demystifying} & 83.26 & 76.52 & 84.97 & 76.18 & 88.30 & 79.34 & 99.77 & 97.70 & 48.81 & 44.38 & 53.31 & 49.04 \\
\app{} & \textbf{85.11} & \textbf{78.84} & \textbf{89.35} & \textbf{81.28} & \textbf{91.66} & \textbf{83.94} & \textbf{99.92} & \textbf{98.89} & \textbf{55.45} & \textbf{49.09} & \textbf{56.97} & \textbf{51.70} \\

\bottomrule
\end{tabular}

}
\caption{\textbf{Invariances of Video representations:} 
\bccomment{The representation learned by \app{} is more invariant to various transformations as compared to baseline MoCo, as shown by the top-k Representation Invariance Score (RIS)~\cite{purushwalkam2020demystifying}. The large improvement in viewpoint invariance is likely due to our strategy of sampling tracked objects with different viewpoints. The large improvement in instance invariance shows that \app{} is better at learning object concepts instead of low-level pixel similarities.  Improved invariance is  useful for object recognition tasks. See Section 4.4 for details of RIS.
}}
\label{tab:invariance}

\end{table*}





%% file: tables/davis.tex
\begin{table}[t]
    \newcommand{\apbbox}[1]{AP$^\text{bbox}_\text{#1}$}
    \newcommand{\apmask}[1]{AP$^\text{mask}_\text{#1}$}
    \newcolumntype{Y}{>{\raggedright\arraybackslash}X}
    \newcolumntype{Z}{>{\centering\arraybackslash}X}

    \centering
    \footnotesize
    \setlength\tabcolsep{1pt}
    \renewcommand{\arraystretch}{1.2}

    \begin{tabular}[]{lcccccc}
        \toprule
        \textbf{Region Similarity $\cal J $} && Mean $\cal M \uparrow$&& Recall $\cal O \uparrow$ & & Decay $\cal D \downarrow$ \\
        \midrule
        \textbf{MoCo} && 0.315 && 0.638  &&  0.025  \\
        \textbf{\app} && 0.544  &&  0.769  &&  -0.014  \\
        \bottomrule
        \end{tabular}
    \caption{\textbf{Unsupervised Tracking on DAVIS 2016: } We show that through our grounding supervision, we are able to better track objects across videos of arbitrary lengths given just the first frame and its associated segmentation map. 
    } 
    \label{tab:davis}
\end{table}

%% file: figures/davis.tex
\begin{figure}[t]
    \centering
    \includegraphics[width=0.48\textwidth]{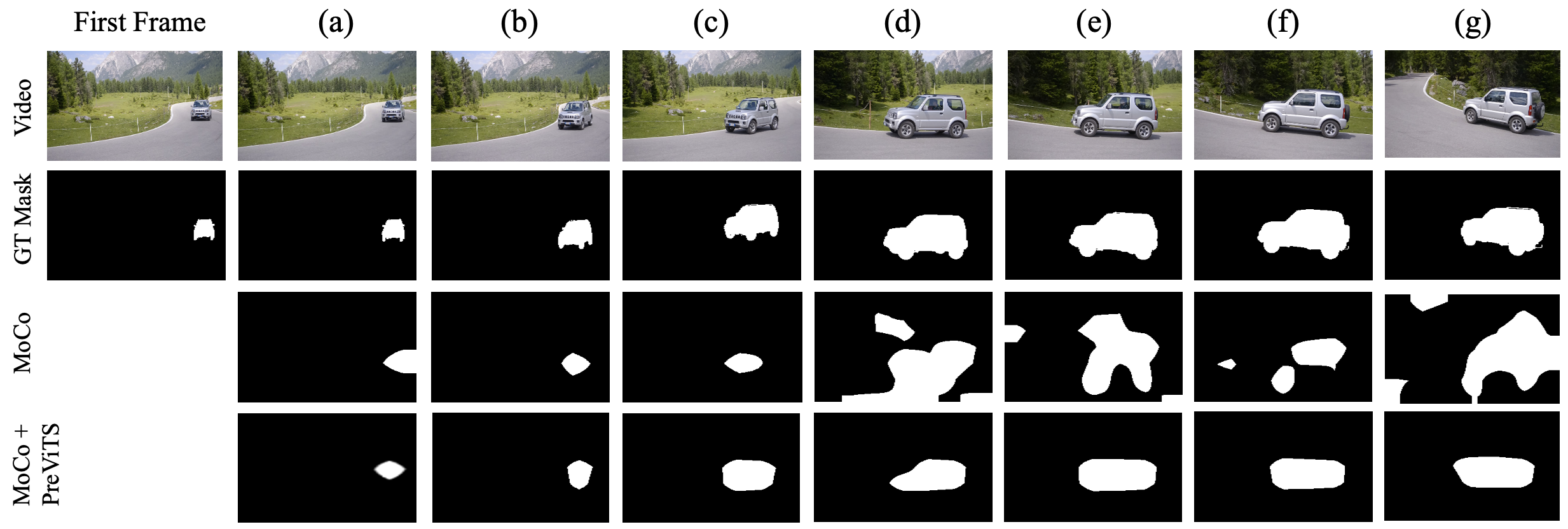}
    \caption{\textbf{Unsupervised Object tracking}. Using Grad-CAM attention and the query-key framework, \app{}-trained model can be used to track objects across the video given the first frame and corresponding segmentation map of the object to track. \app{} is able to localize objects under viewpoint changes, while the baseline model is unable to do so.}
    \label{fig:davis}
\end{figure}

%% file: tables/ablation.tex
\begin{table}[t]
    \centering \footnotesize
    \setlength{\tabcolsep}{4pt}

    \begin{subfigure}[t]{\linewidth}
        \newcommand{\apbbox}[1]{AP$^\text{bbox}_\text{#1}$}
        \newcommand{\apmask}[1]{AP$^\text{mask}_\text{#1}$}
        \newcolumntype{Y}{>{\raggedright\arraybackslash}X}
        \newcolumntype{Z}{>{\centering\arraybackslash}X}
    
        \centering
        \footnotesize
        \setlength\tabcolsep{1pt}
        \renewcommand{\arraystretch}{1.1}
    \begin{tabularx}{\linewidth}{l@{\hskip 0.6in}c@{\hskip 0.2in}c@{\hskip 0.2in}c}
    
        \toprule
        \textbf{Temporal Sampling} & Varying $\delta$ & Constant $\delta$ & $\delta = 0$ \\
        \midrule
        \textbf{\vocclf{}} & \graycell 73.0 & 72.4\Drop{0.6} &  67.5\Drop{5.5} \\
        \textbf{\ucf{}} & \graycell 84.5 & 83.7\Drop{1.8} & 84.3\drop{0.2} \\
        \bottomrule
        \end{tabularx}
        
        \vspace{2pt}
        \caption{\small Effect of different temporal sampling strategy.}
        \label{subtab:ablations1}
        \vspace{10pt}
    \end{subfigure}
    
    \begin{subfigure}[t]{\linewidth}
        \begin{tabularx}{\linewidth}{lXcccc}
        \toprule
        \textbf{Spatial area threshold} && $\mu= 0.0$& $\mu= 0.2$& $\mu= 0.3$ & $\mu = 0.4$ \\
        \midrule
        \textbf{\vocclf{}} && 71.5\Drop{1.5} & 72.1\Drop{0.9} &  \graycell 73.0  & 72.8\drop{0.2} \\
        \textbf{\ucf{}} && 83.7\Drop{3.7} & 85.1\Rise{0.6} & \graycell 84.5  & 84.2\drop{0.3} \\
        \bottomrule
        \end{tabularx}
        \vspace{2pt}
        \caption{\small Effect of Area threshold $\mu$ (Fixing $\mu = 0.3$)}
        \label{subtab:ablations2}
        \vspace{10pt}
    \end{subfigure}

    \begin{subfigure}[t]{\linewidth}
        \begin{tabularx}{\linewidth}{lXcccc}
        \toprule
        \textbf{Loss weighing factor} && $\lambda = 0.0$ & $\lambda = 2.0$ & $\lambda = 3.0$ & $\lambda = 4.0$ \\
        \midrule
        \textbf{\vocclf{}} && 70.3\Drop{2.7} & 72.4\Drop{0.6} &  \graycell73.0 & 72.6\drop{0.4} \\
        \textbf{\ucf{}} &&    80.8\Drop{3.7} & 83.4\Drop{2.1} &  \graycell84.5 & 84.1 \Drop{0.6} \\
        \bottomrule
        \end{tabularx}
        \vspace{2pt}
        \caption{\small Effect of loss weighing factor $\lambda$ (Fixing $\lambda = 3.0$)}
        \label{subtab:ablations3}
        \vspace{10pt}
    \end{subfigure}

    \begin{subfigure}[t]{\linewidth}
        \begin{tabularx}{\linewidth}{lXcccc}
        \toprule
        \textbf{Tracking} &&No Tracking& Unsup. Box & Unsup. Mask & Sup. Seg \\
        \midrule
        \textbf{\vocclf{}} && 68.3\Drop{4.7} & 71.9\Drop{1.1} &\graycell 73.0 &  75.0 \Rise{2.0} \\
        \textbf{\ucf{}} && 79.0  \Drop{5.5} & 83.0\Drop{1.5} &\graycell 84.5  & 86.1 \Rise{1.6} \\
        \bottomrule
        \end{tabularx}
        \vspace{2pt}
        \caption{\small Effect of different tracking supervision}
        \label{subtab:ablations4}
        \vspace{10pt}
    \end{subfigure}
    
    
    \caption{
        \textbf{Ablations for \app{} training:}
        We isolate the effects of our training components. We find that
         \textbf{(a)} starting with a shorter temporal distance between query-key clips and relaxing the constraint as training progresses improves performance. 
         \textbf{(b)} adding some amount of spatial constraints based on IoU with tracking mask ensures that different clips contain common salient regions and this improves performance.
        \textbf{(c)} increasing weights on attention loss increases the downstream performance up to a certain point,
        \textbf{(d)} \bccomment{replacing unsupervised video tracking supervision with a noisy bounding box tracking tube  achieved a significant gain over the baseline. Apply supervised tracking improves downstream performance slightly.}
    }
    \label{tab:train_ablations}

\end{table}

%% file: figures/visualization.tex
\begin{figure*}[t]
    \centering
    \includegraphics[width=\textwidth]{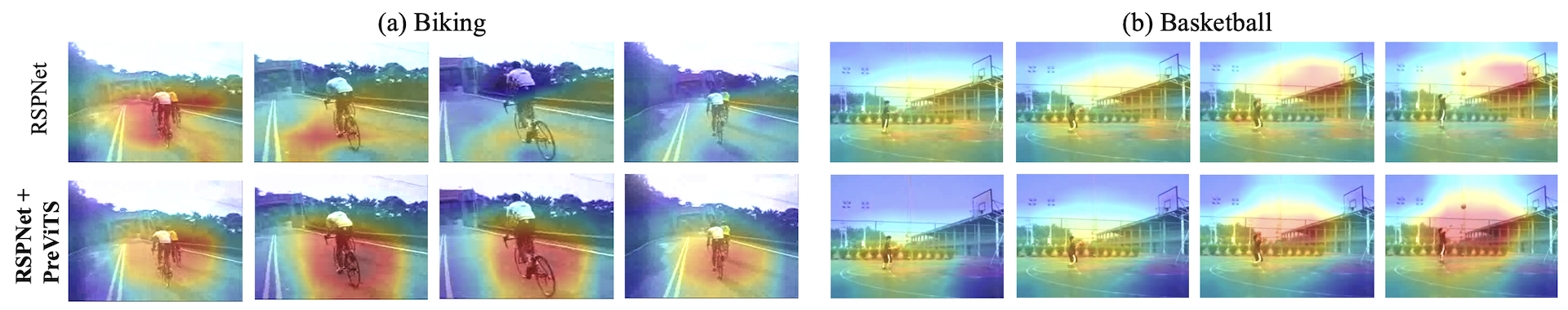}
    \caption{\textbf{Visual Grounding for Action Classification.} \app{} provides better visual grounding as shown by Grad-CAM attention maps of pretrained models finetuned on UCF-101.   In (a), our model focuses on the human and bike while the baseline model attends to seemingly irrelevant regions, including the road in the background. In (b), our model attends to the man and the ball in the air in addition to the basketball court while the baseline model focuses mostly on the court.}
    \label{fig:ucf}
\end{figure*}

%% file: figures/occlusion.tex
\begin{figure}[t]
    \centering
    \includegraphics[width=0.48\textwidth]{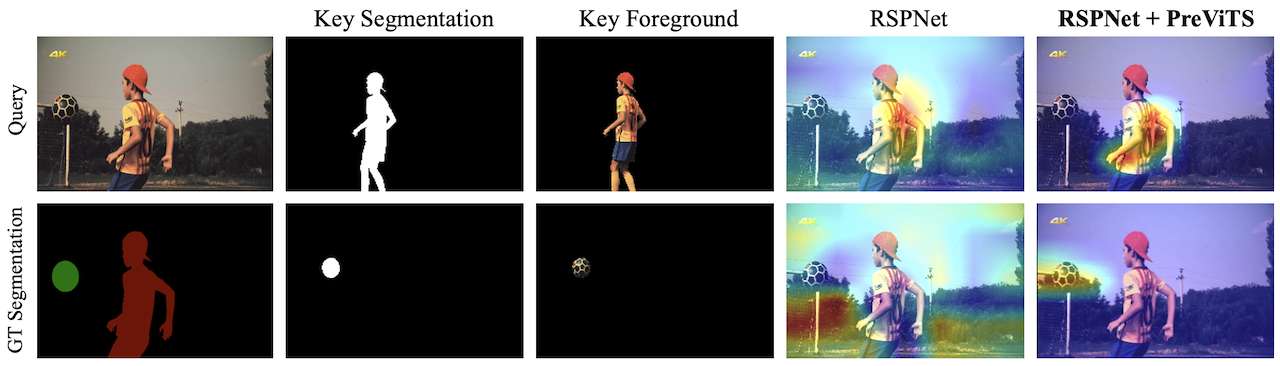}
    \caption{\textbf{Discriminative localization of objects.} When provided query with two different segmentation corresponding to different foreground objects and key foregrounds, \app{}-trained model is able to localize the object accurately, capturing class-specific semantic discrimination between objects.} 
    \label{fig:occlusion}
\end{figure}

%% file: sections/5_conclusion.tex
\section{Conclusion}
\paragraph{Limitations and potential impact:}
Our method has a few limitations. First, acquiring and utilizing unsupervised tracking requires additional computational resources. Also, since our current tracking method captures the most salient object in the video, we do not model multi-object interaction in the video, which is an interesting future work direction. Moreover, our pretraining datasets are relatively cleaner than random videos on YouTube. It is unknown if our method can generalize to the different genres such as news and gaming. Finally, our pretraining datasets may contain unintended societal, gender, racial, and other biases, whose effect was not examined in the current work.

\paragraph{Concluding remarks:} 
We propose a visual self-supervised network that learns to localize foreground objects present in video data utilizing unsupervised tracking supervision. Experiments on various video downstream tasks show that guiding the model to focus on the foreground region is beneficial for accurate video representations self-supervised learning. Also, we demonstrate different properties of our learned features, which capture viewpoint, occlusion, illumination, and instance invariances. The result of our model shows better grounding ability with less background bias. We hope that our method leads to further research on robust, accurate and grounded visual representation learning from large-scale uncurated video data from the internet.

%% file: tables/image_ex.tex
\begin{table*}[ht]
    \newcommand{\apbbox}[1]{AP$^\text{bbox}_\text{#1}$}
    \newcommand{\apmask}[1]{AP$^\text{mask}_\text{#1}$}
    \newcolumntype{Y}{>{\raggedright\arraybackslash}X}
    \newcolumntype{Z}{>{\centering\arraybackslash}X}

    \centering
    \footnotesize
    \setlength\tabcolsep{1pt}
    \renewcommand{\arraystretch}{1.2}

    \begin{tabularx}{\textwidth}{c l c c c c c c c YYY c YYYYYY}
    \toprule
    ~
    & \multicolumn{1}{l}{\bf \multirow[b]{2}{*}{Method}} &&  \multicolumn{1}{l}{\bf \multirow[b]{2}{*}{Dataset}}
    & ~~~
    & \multicolumn{1}{c}{\bf \vocclf{} clf.}
    &~~& \multicolumn{1}{c}{\bf \inclf{} clf.}
    &~~& \multicolumn{3}{c}{\bf \voc{} Detection}
    &~~& \multicolumn{6}{c}{\bf COCO Instance Segmentation} \\
    
    \cmidrule{6-6} \cmidrule{8-8} \cmidrule{10-12} \cmidrule{14-19}
    
    ~ & ~ & ~  & && \multicolumn{1}{c}{mAP} && \multicolumn{1}{c}{Top-1 acc.} &&
            \apbbox{all} & \apbbox{50} & \apbbox{75} &&
            \apbbox{all} & \apbbox{50} & \apbbox{75} &
            \apmask{all} & \apmask{50} & \apmask{75} \\

        \band
    \midrule
    \ttbf{1)}  & \random{} &&  &&
                -- && -- &&
                33.8 & 60.2 & 33.1 &&
                36.7 & 56.7 & 40.0 & 33.7 & 53.8 & 35.9 \\
    \ttbf{2)}  & ImageNet Fully Sup &&  &&
                -- && -- &&
                53.5 & 81.3 & 59.1 && 
                38.9 & 59.6 & 42.7 & 35.4 & 56.5 & 38.1 \\
            \midrule    
    \ttbf{3)}  & MoCo && K400 &&
                69.3 && 47.3 &&                             
                50.6 & 78.0 & 55.1 &&                       
                40.5 & 58.9 & 41.9 & 35.1 & 55.6  & 37.3 \\  
    
    \ttbf{4)}  & \hspace{5pt} + Tracking Con. Sampling && K400 &&
                70.4\Rise{1.1} && 48.2\Rise{0.9} &&                             
                51.2\Rise{0.6} & 78.4\Rise{0.4} & 56.1\Rise{1.0} &&                       
                40.8\Rise{0.3} & 59.5\Rise{0.6} & 42.6\Rise{0.7} & 35.8\Rise{0.7} & 56.8\Rise{1.2} & 38.3\Rise{1.0} \\  
    
    \ttbf{5)}  & \hspace{5pt} + \app{} && K400 &&
                71.2\Rise{1.9} && 48.6\Rise{1.3} &&                             
                51.8\Rise{1.2} & 78.3\rise{0.3} & 56.0\Rise{0.9} &&                       
                41.0\Rise{0.5} & 59.4\rise{0.5} & 42.8\Rise{0.9} & 35.6\Rise{0.5} & 57.2\Rise{1.6} & 38.4\Rise{1.1} \\  
       \midrule    
    \ttbf{6)}  & MoCo  && {\scriptsize VGG Sound} &&
                68.3 && 46.9 &&                              
                48.3 & 76.5 & 52.6 &&                       
                38.4 & 58.7 & 41.9 & 35.0 & 55.8  & 37.2 \\  
    
    \ttbf{7)}  & \hspace{5pt} + Tracking Con. Sampling && {\scriptsize VGG Sound} &&
                70.3\Rise{2} && 48.1\Rise{1.2} &&                             
                49.0\Rise{0.7} & 77.1\Rise{0.6} & 52.7\rise{0.1} &&                       
                38.3\drop{0.1} & 58.7\rise{0.0} & 41.7\drop{0.2} & 35.0\rise{0.0} & 55.9\rise{0.1} & 37.6\Rise{0.4} \\  
    
    \ttbf{8)}  & \hspace{5pt} + \app{} && {\scriptsize VGG Sound} &&
                73.0\Rise{4.7} && 50.6\Rise{3.7} &&                             
                52.5\Rise{4.2} & 78.7\Rise{2.2} & 55.1\Rise{2.5} &&                       
                39.4\Rise{1.0} & 59.8\Rise{1.1} & 43.0\Rise{1.1} & 35.7\Rise{0.7} & 56.8\Rise{1.0} & 38.2\Rise{1.0} \\  

    \bottomrule
    \end{tabularx}
    \vspace{-5pt}
    \caption{\textbf{Transfer Learning on Image Downstream Tasks:}
    On tasks using linear probes (VOC and ImageNet classification) and finetuning (VOC Detection, COCO Segmentation), \app{} outperforms baseline MoCo when evaluated on models pretrained on VGG-Sound and Kinetics-400. 
    We color the difference $\ge 0.5$ to show improvement over the baseline MoCo models (row 3 and 6).
    } 
    \label{tab:main_results}
\end{table*}

%% file: sup/percentage.tex
\begin{figure}[th]
    \centering
    \includegraphics[width=0.5\textwidth]{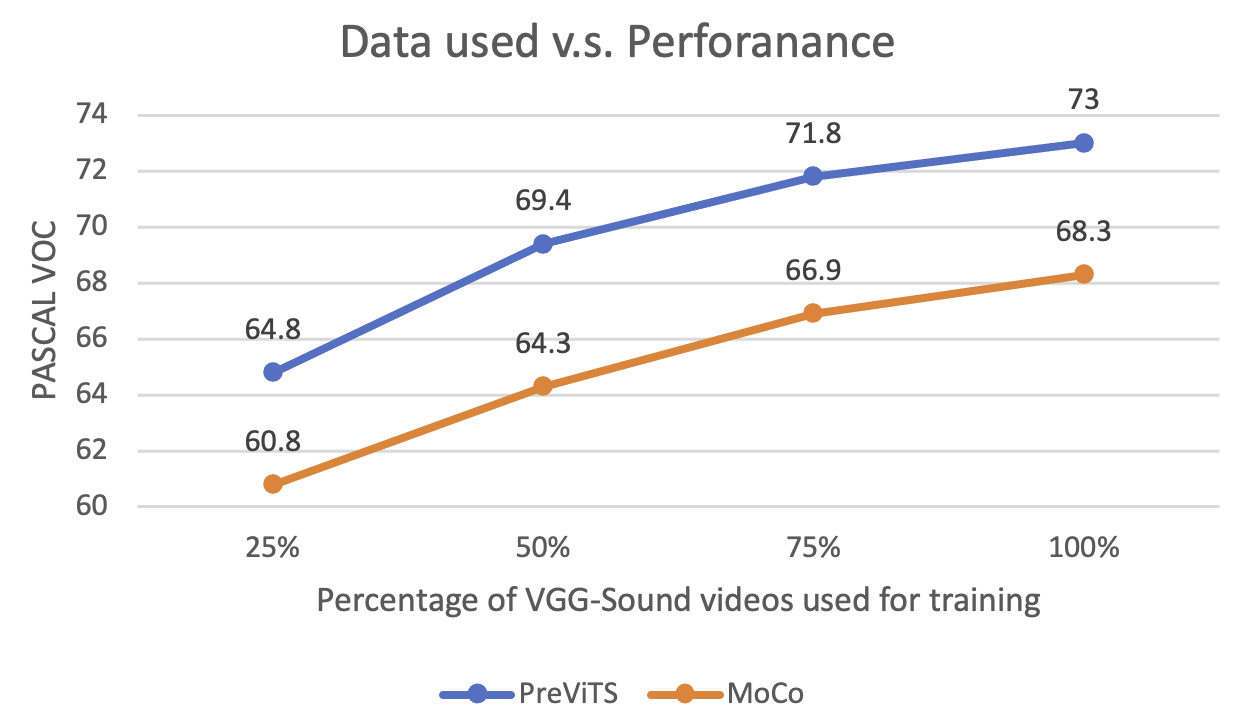}
    \caption{Percentage of VGG-Sound videos used for training.}
    \label{fig:percentage}
\end{figure}

%% file: figures/image_back.tex
\begin{figure}[t]
    \centering
    \includegraphics[width=0.5\textwidth]{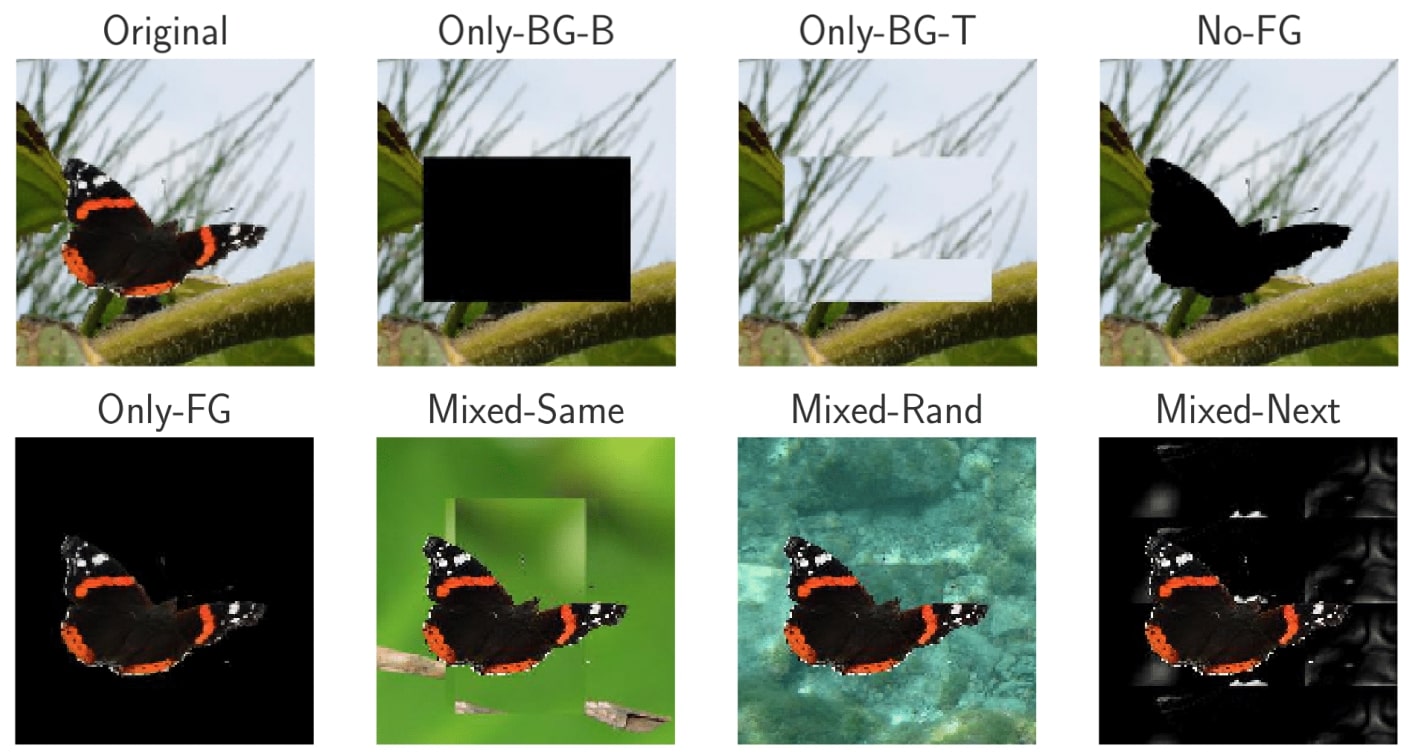}
    \caption{Image Background Challenge Settings}
    \label{fig:backgrounds}

\end{figure}

%% file: sup/sup_ucf.tex
\begin{figure*}[th]
    \centering
    \includegraphics[width=0.9\textwidth]{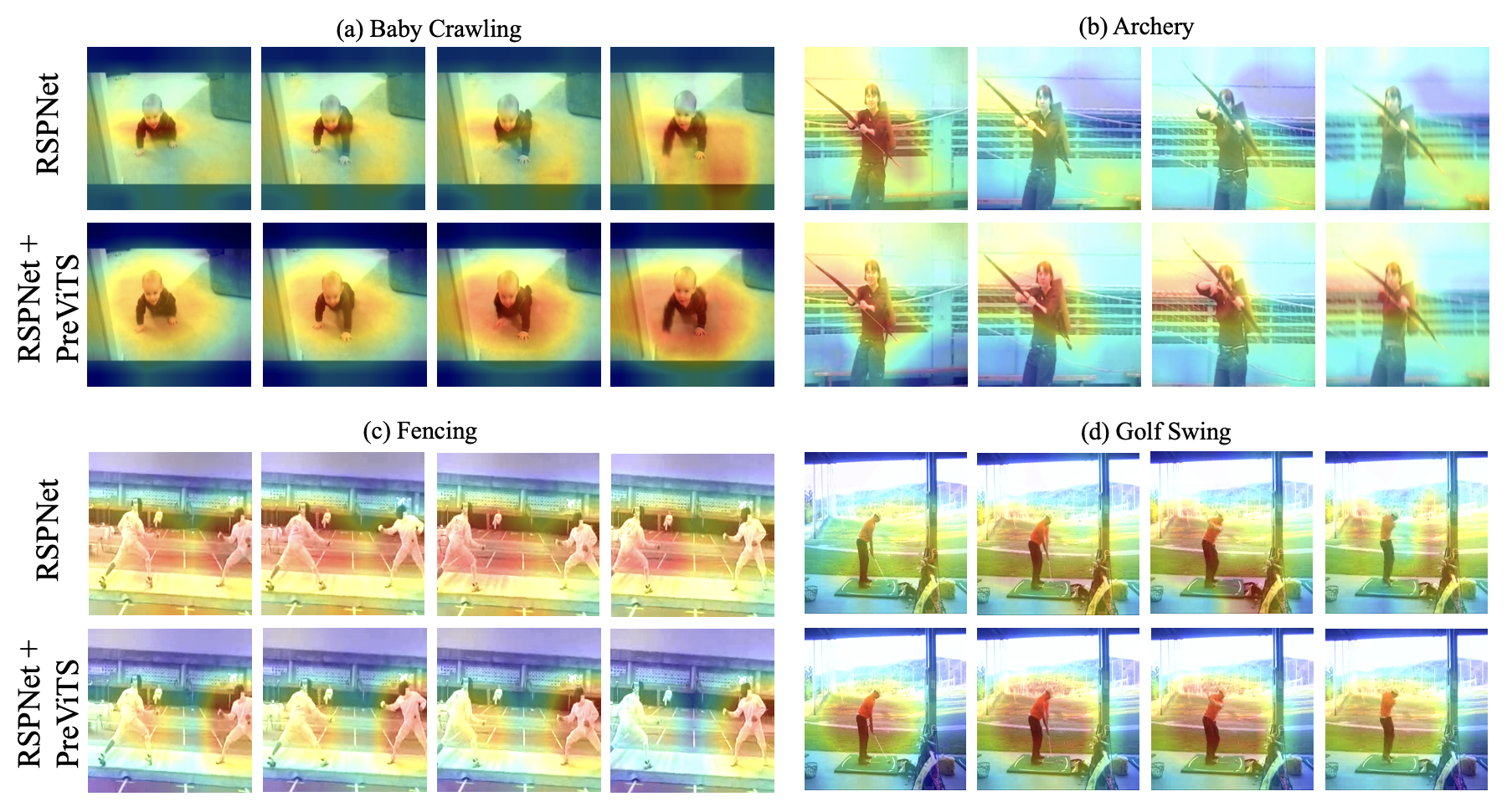}
    \caption{Grad-CAM Visualization for UCF-101 Action Classification. }
    \label{fig:sup_ucf}

\end{figure*}

%% file: sup/sup_bg.tex
\begin{figure*}[t]
    \centering
    \includegraphics[width=0.9\textwidth]{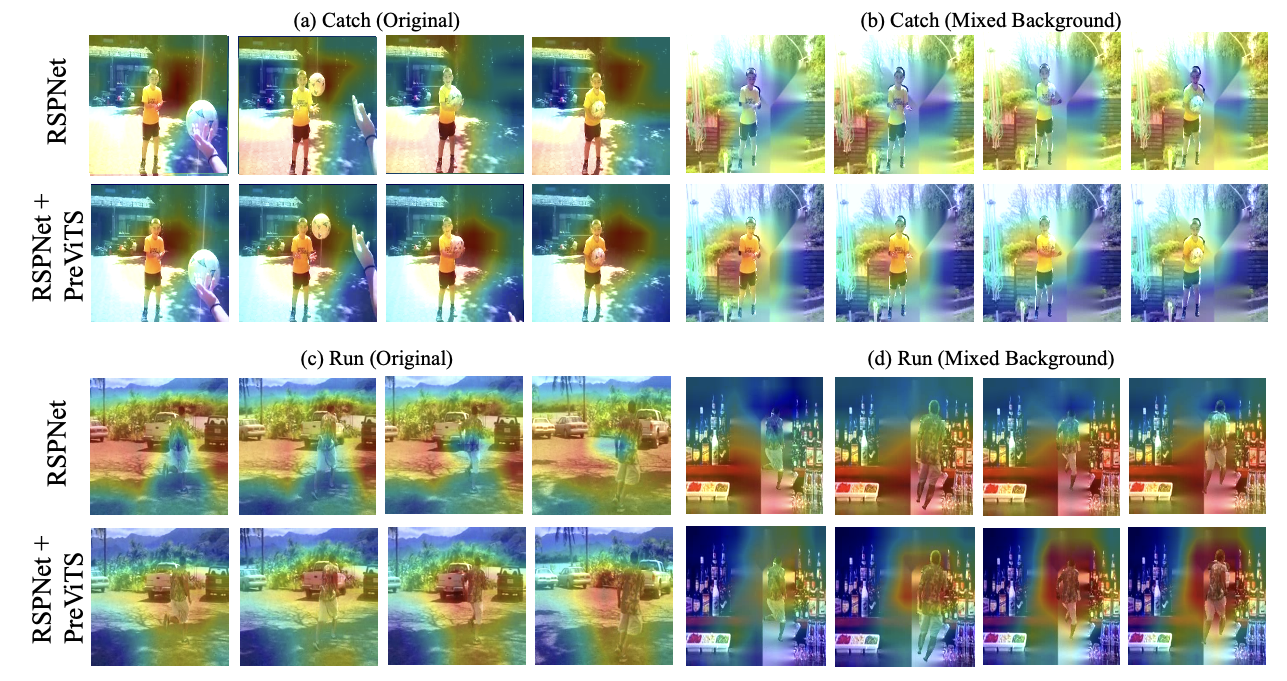}
    \caption{Grad-CAM Visualization for Video Backgrounds Challenge. }
    \label{fig:sup_bg}

\end{figure*}

%% file: sup/sup_davis.tex
\begin{figure*}[th]
    \centering
    \includegraphics[width=\textwidth]{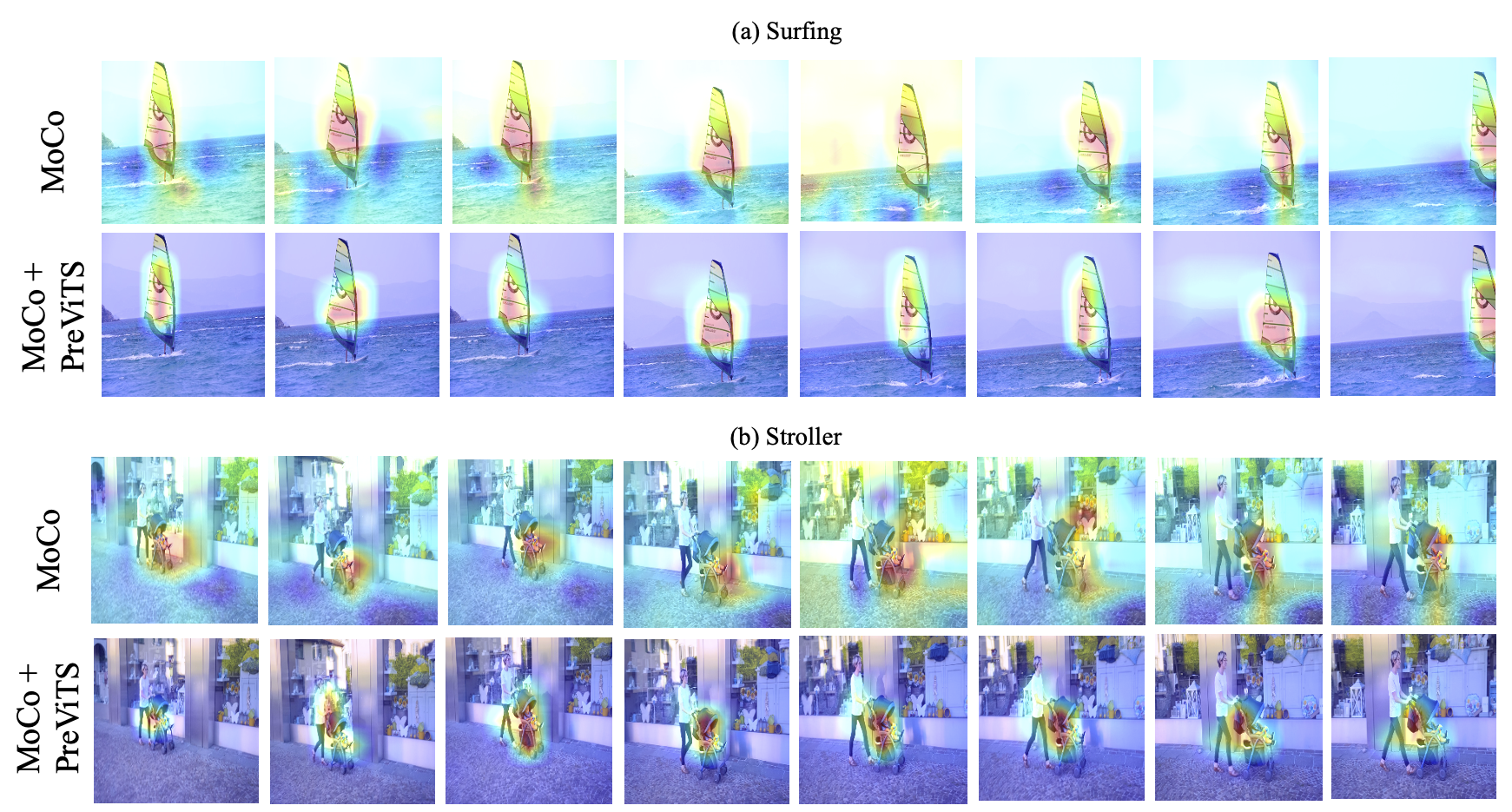}
    \caption{Grad-CAM Visualization for DAVIS Video Object Tracking and Segmentation. }
    \label{fig:sup_davis}

\end{figure*}

%% file: sup/sup_davis2.tex
\begin{figure*}[th]
    \centering
    \includegraphics[width=\textwidth]{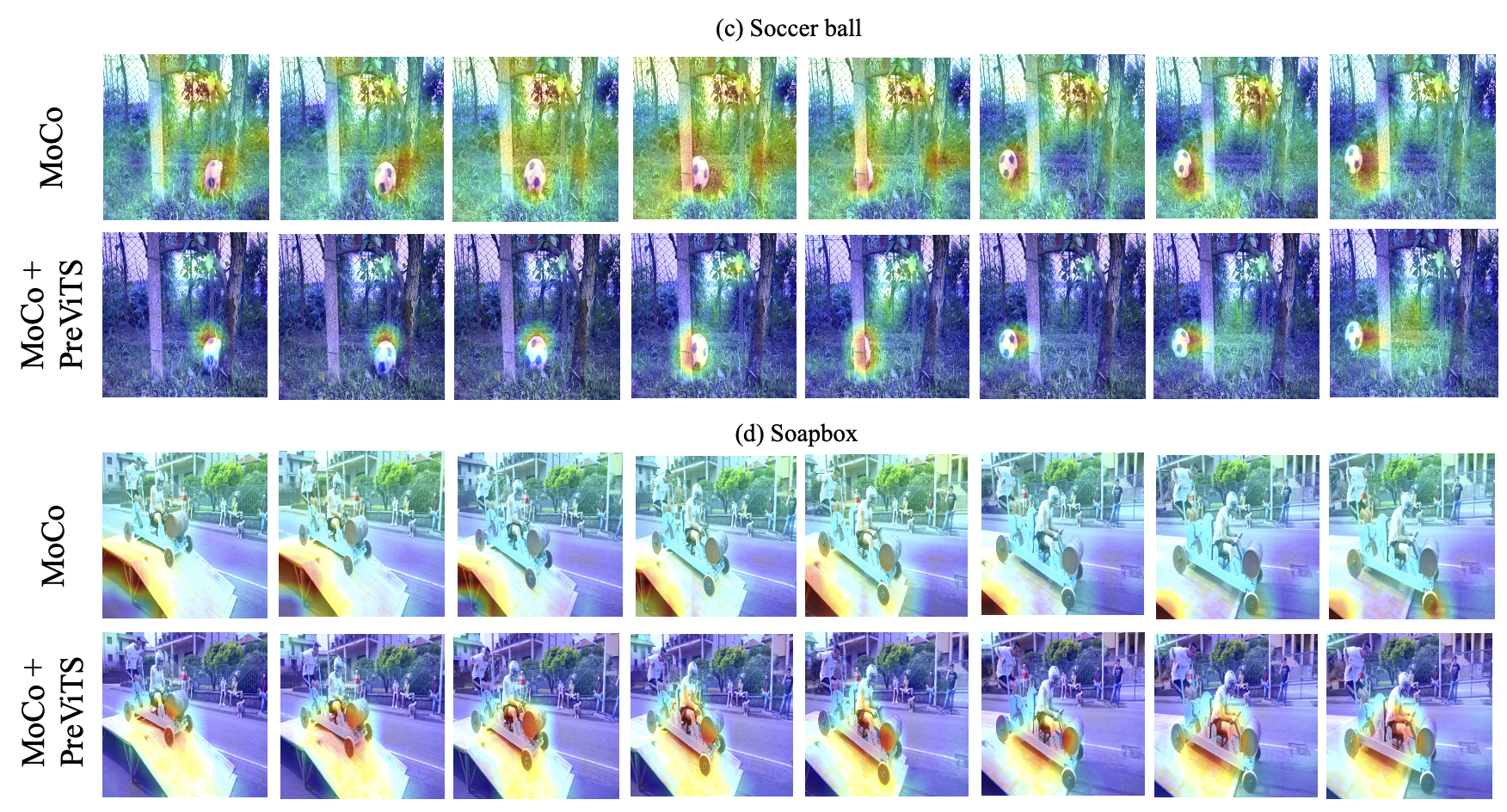}
    \caption{Grad-CAM Visualization for DAVIS Video Object Tracking and Segmentation. }
    \label{fig:sup_davis2}

\end{figure*}